\documentclass{article}

\PassOptionsToPackage{numbers,compress}{natbib}
\usepackage[preprint]{neurips_2026}

\usepackage[utf8]{inputenc}
\usepackage[T1]{fontenc}
\usepackage{hyperref}
\usepackage{url}
\usepackage{booktabs}
\usepackage{amsfonts}
\usepackage{amsmath}
\usepackage{nicefrac}
\usepackage{microtype}
\usepackage{xcolor}
\usepackage{wrapfig}
\usepackage{graphicx}
\usepackage{amssymb}

\title{Evolving-RL: \\ End-to-End Optimization of Experience-Driven Self-Evolving Capability within Agents}

\author{%
  \textbf{Zhiyuan Fan}$^{1,2}$, 
  \textbf{Wenwei Jin}$^{1,\dagger}$, 
  \textbf{Feng Zhang}$^{1}$, 
  \textbf{Bin Li}$^{1}$, 
  \textbf{Yihong Dong}$^{2}$, 
  \textbf{Yao Hu}$^{1}$, 
  \textbf{Jiawei Li}$^{1}$ \\[0.4em]
  $^{1}$Xiaohongshu Inc.\quad $^{2}$School of Computer Science, Peking University \\[0.2em]
  $^{\dagger}$Corresponding author. \\[0.2em]
  \texttt{\{zyfan043, wenwei1217.jin, libin656712945, yaoohu\}@gmail.com} \\
  \texttt{\{zhangfeng4, wangdesheng\}@xiaohongshu.com} \\
  \texttt{dongyh@stu.pku.edu.cn}
}

\begin{document}

\maketitle

\begin{abstract}

Experience-driven self-evolving agents aim to overcome the static nature of large language models by distilling reusable experience from past interactions, thus enabling adaptation to novel tasks at deployment time. This process places substantial demands on the foundation model’s capacities for abstraction, generalization, and in-context learning. However, most existing studies focus primarily on system-level design choices, such as how experience is represented and managed, neglecting the inherent capabilities of the underlying model. While some recent works have started to optimize the experience utilization stage via reinforcement learning, they still fail to treat self-evolution as a unified process to be jointly optimized. To this end, we propose \textbf{Evolving-RL}, an efficient algorithmic framework that jointly improves the experience extraction and utilization capabilities required for self-evolution. Specifically, we center the learning process on experience extraction and evaluation, using the two supervisory signals derived from evaluation to optimize the extractor and solver separately and thus enable their coordinated co-evolution. Experiments on ALFWorld and Mind2Web show that Evolving-RL effectively enhances LLMs’ ability to extract and reuse experience, leading to strong performance gains on out-of-distribution tasks (up to 98.7\% relative improvement over the GRPO baseline on ALFWorld unseen tasks and 35.8\% on Mind2Web), and these gains are fully unlocked only through the coordinated co-evolution of experience extraction and utilization. Furthermore, Evolving-RL inherently functions as an experience-augmented RL algorithm. By internalizing reusable experience patterns directly into model parameters, it achieves remarkable performance gains over standard baselines on both seen and unseen tasks, even in the absence of test-time experience accumulation. Our code is available at \url{https://github.com/Fanzy27/Evolving-RL}.

\end{abstract}

\section{Introduction}
\label{sec:intro}

Large language models (LLMs) have demonstrated remarkable capabilities across a broad range of tasks, including complex reasoning~\cite{huang-chang-2023-towards, qiao-etal-2023-reasoning, kojima2022large, wei2022chain} and autonomous agent decision-making~\cite{wang2024survey, guo2024multiagents, yao2023react}. However, once trained, LLMs are largely static: they lack the ability to continually adapt themselves to the complex out-of-distribution environments and tasks encountered during deployment. This fundamental limitation has motivated a growing body of research into test-time self-evolution~\cite{gao2026surveyselfevolvingagentswhat, xiang2026systematic, shinn2023reflexion, zhao2023expel, agrawal2026gepareflectivepromptevolution}. As a prominent direction within this paradigm, experience-driven self-evolving agents accumulate reusable experiences from prior interactions and leverage them to solve related future tasks, thereby progressively enhancing their deployment-time capabilities.

\begin{figure}[t]
    \centering
    \includegraphics[width=\linewidth]{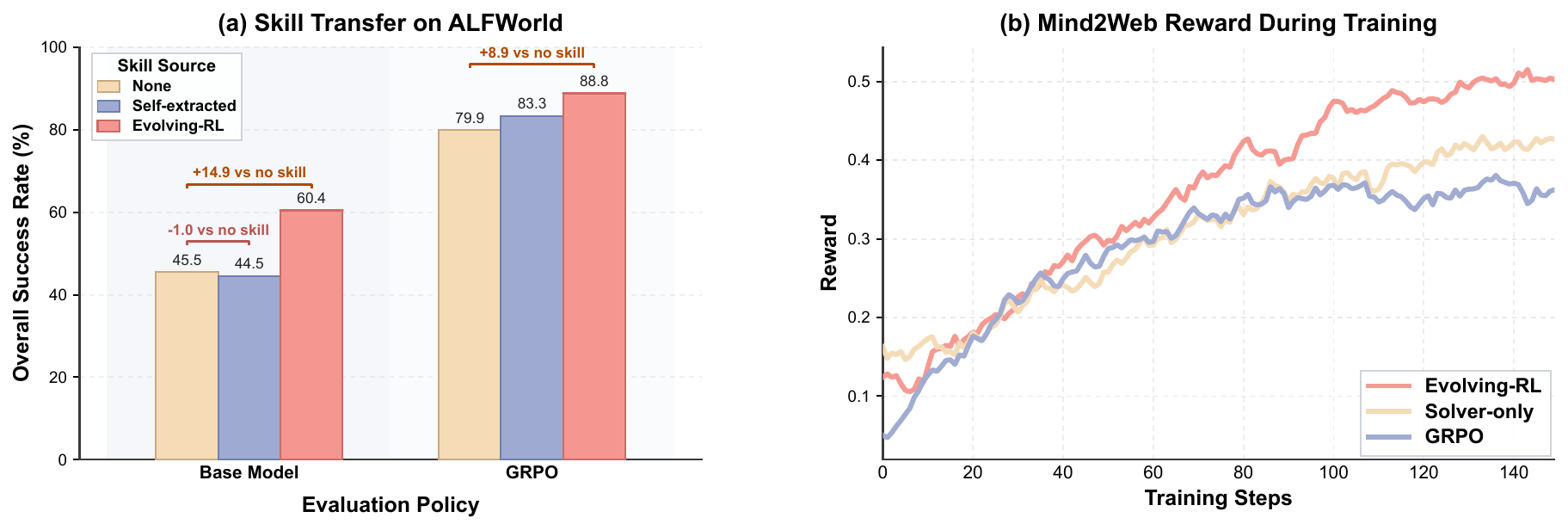}
    \caption{
    (a) Skills accumulated by our method (``Evolving-RL'') transfer effectively to other policies and consistently improve downstream performance. Conversely, hindered by inferior skill quality, skills extracted by the base model itself (``self-extracted'') not only fail to improve, but actually degrade performance compared to no injection (``None''), demonstrating the critical role of our method in ensuring skill quality.
    (b) Beyond test-time self-evolution, Evolving-RL proves highly effective as an experience-augmented RL algorithm, consistently outperforming both standard GRPO and solver-only training that is constrained by a static skill library.
    }
    \label{fig:figure1}
\end{figure}

Existing experience-driven self-evolution approaches primarily focus on system-level evolutionary design, with considerable research devoted to experience representation~\cite{yang2024copsempoweringllm, zheng2025skillweaverwebagentsselfimprove, wang2024agentworkflowmemory, ouyang2026reasoningbank}, formulation, and management mechanisms~\cite{cao2026remembermerefineme,  tang2025agentkbleveraging, yang2025learningjobexperiencedrivenselfevolving}. While these prompt-based methods have successfully demonstrated that injecting past experiences can significantly enhance downstream decision-making, their effectiveness is ultimately bounded by the underlying model’s ability to extract and leverage these experiences~\cite{shao2026agentmisevolveemergentrisks}—a process that heavily relies on the model possessing robust in-context learning~\cite{brown2020language} and abstract reasoning capabilities.


Several recent studies~\cite{xia2026skillrl, wu2025evolverselfevolvingllmagents} have explored reinforcement learning as a way to enhance the model’s ability to utilize experience. However, these methods do not optimize self-evolution as a unified process. They improve only the utilization phase, while relying on stronger external models or hand-crafted filtering mechanisms to ensure the quality of extracted experience. Such a decoupled design is inherently limited, as extraction and utilization are not merely complementary capabilities, but mutually constitutive during learning. The quality of extracted experience directly influences the reinforcement learning dynamics for experience utilization. When the provided experience is noisy, conditioning on it introduces variance and ambiguity into the optimization process. Consequently, the policy often converges to a degenerate behavior: it learns to systematically ignore the provided experience, as bypassing unreliable guidance yields more stable returns than attempting to exploit it.

To bridge this gap, we propose \textbf{Evolving-RL}, an efficient algorithmic framework that jointly optimizes both experience extraction and utilization within a unified training paradigm. To make this joint optimization tractable and interpretable, we specifically instantiate experience as textual skills—compact procedural abstractions that capture actionable regularities regarding what to do, when to intervene, and how to recover from failures. Anchored by this representation, Evolving-RL employs an extractor-centric design: for each source interaction, a shared policy generates candidate textual skills. Each candidate is then rigorously evaluated by applying it to multiple retrieved instances and receiving reward according to downstream feedback. Concurrently, the trajectories collected during this downstream evaluation are recycled to refine the model's skill-utilization capability. By utilizing downstream transfer performance as the primary supervisory signal, extractor and solver co-evolve, naturally coupling advancements in transferable skill abstraction with robust skill-conditioned execution.


We evaluate our method on ALFWorld, a text-based embodied task environment, and Mind2Web, a web-based task benchmark. Empirical results show that Evolving-RL significantly improves the model’s ability to explicitly extract and reuse skills, thereby leading to markedly better generalization on out-of-distribution tasks. Specifically, when augmented with extracted skills, Evolving-RL boosts the success rate on ALFWorld unseen tasks from 44.6\% (the GRPO~\cite{shao2024deepseekmath} baseline) to 88.6\%, and increases overall action accuracy on Mind2Web from 22.73\% to 30.87\%. Moreover, in the skill transfer experiments, the high-quality skills extracted by Evolving-RL-trained models exhibit strong transferability to other models (Figure~\ref{fig:figure1}a). In addition, Evolving-RL also strengthens the underlying policy itself by internalizing reusable experience patterns into the model parameters. Even without test-time skill injection, our trained policy still achieves an 81.1\% success rate on ALFWorld unseen tasks and a 28.05\% overall action accuracy on Mind2Web, substantially outperforming the standard GRPO baselines (33.7\% and 22.83\%, respectively), demonstrating its value not only for self-evolution, but also as a form of experience-augmented reinforcement learning (Figure~\ref{fig:figure1}b).







\section{Related Work}
\label{sec:related}

\subsection{Experience-Driven Self-Evolving Agent}
 Experience-driven self-evolution enables agents to continuously accumulate and reuse knowledge, overcoming the static nature of their post-training capabilities \cite{zhao2023expel, gao2026surveyselfevolvingagentswhat}. Existing literature in this domain has advanced this paradigm across multiple dimensions. In terms of experience representation, prior studies have explored diverse granularities, ranging from raw interaction trajectories~\cite{zhou2025memento, yang2024copsempoweringllm}, reusable workflows~\cite{wang2024agentworkflowmemory} to executable skills~\cite{zheng2025skillweaverwebagentsselfimprove, alzubi2026evoskillautomatedskilldiscovery, zhou2026mementoskillsletagentsdesign}, strategic principles~\cite{ouyang2026reasoningbank}. Beyond representation, parallel efforts have also developed sophisticated mechanisms for autonomous experience extraction~\cite{alzubi2026evoskillautomatedskilldiscovery, ma2026skillclawletskillsevolve, shen2026skillfoundrybuildingselfevolvingagent} and efficient memory management~\cite{tang2025agentkbleveraging, cai2025flex, yang2025learningjobexperiencedrivenselfevolving}. Despite these advancements, these systems predominantly rely on the base model's pre-existing ability to extract and utilize effective experience. Consequently, the benefits of experience-driven evolution are fundamentally bounded by the underlying model's intrinsic capacities~\cite{shao2026agentmisevolveemergentrisks}. While recent methods~\cite{wu2025evolverselfevolvingllmagents, xia2026skillrl, tu2026dynamicdualgranularityskillbank} attempt to employ Reinforcement Learning (RL) to enhance the underlying model's capabilities, they typically optimize experience utilization in isolation, failing to treat extraction and utilization as a unified process for joint optimization. Instead, their experience extraction phases remain heavily dependent on hand-crafted filtering heuristics or more capable external models. Motivated by this limitation, our work optimizes the entire self-evolution loop: rather than treating experience generation as an external component, we drive the end-to-end evolution process based on the transfer value of the extracted skills.

 \subsection{Experience-Augmented Reinforcement Learning}
 Enhancing experience utilization through reinforcement learning (RL) inherently constitutes an experience-augmented RL paradigm~\cite{shi2026experientialreinforcementlearning, xia2026skillrl, li2026ariseagentreasoningintrinsic}, which can significantly boost training efficiency. Therefore, some recent works also define this process as an RL paradigm. ERL~\cite{shi2026experientialreinforcementlearning} and RetroAgent~\cite{zhang2026retroagentsolvingevolvingretrospective} leverage task-specific reflections as experience to augment the learning process. Although yielding effective within-task gains, such experience lacks generalizability and is difficult to transfer across tasks. Alternatively, frameworks like EvolveR~\cite{wu2025evolverselfevolvingllmagents}, SkillRL~\cite{xia2026skillrl} and D2Skill~\cite{tu2026dynamicdualgranularityskillbank} distill trajectories into reusable experience libraries that are consumed during RL training. However, their experience-construction pipelines remain weakly optimized, as their extraction processes still rely heavily on manually crafted filtering mechanisms or stronger external models. In contrast, Evolving-RL treats experience extraction and utilization as a unified optimization problem. We instantiate experience as transferable textual skills, evaluate them by their impact on retrieved downstream tasks, and leverage the resulting feedback to co-train the extractor and the solver within a single RL loop, thereby ensuring that the extracted experience consistently provides effective utility for RL training.
 
Notably, a concurrent work~\cite{muhtar2026complementaryreinforcementlearning} shares a similar motivation of co-optimizing experience extraction and utilization. However, they erroneously attribute the training instability during co-evolution to parameter conflicts, ultimately training the two capabilities separately across two distinct models. Furthermore, their experience validation remains confined to single-task scenarios, which limits cross-task generalizability. In contrast, Evolving-RL explicitly ensures generalizability through evaluation-centric training, and as we detail in Appendix~\ref{app:a1}, effectively resolves the underlying instability to train both capabilities within a truly unified, single model.

\section{Evolving-RL}
\label{sec:method}

\subsection{Framework Overview}

\begin{figure}[t]
    \centering
    \includegraphics[width=\linewidth]{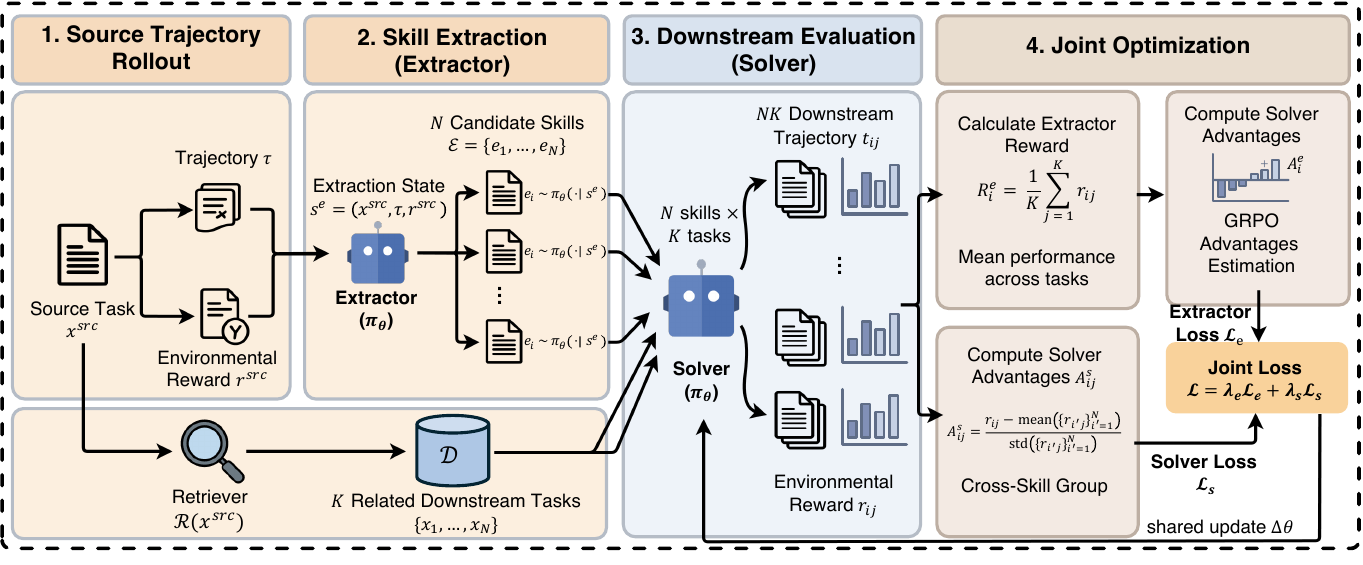}
    \caption{Overview of Evolving-RL. The framework begins with online skill extraction, followed by downstream evaluation of the extracted skills. This evaluation process produces training signals for both the extractor and the solver, enabling their joint optimization within a unified co-evolutionary framework.}
    \label{fig:overview}
\end{figure}

We propose \textbf{Evolving-RL}, a unified algorithmic framework that jointly improves the model's skill extraction and skill utilization capabilities within a single shared policy. As illustrated in Figure~\ref{fig:overview}, our method generates skills online and evaluates them on downstream tasks. This process produces two coupled learning signals. Aggregated downstream rewards across multiple tasks provide direct supervision on the quality and generality of the extracted skills, thereby improving the model's skill extraction capability. Meanwhile, the skill-conditioned trajectories collected during downstream evaluation are reused to improve its skill utilization capability. In this way, all sampled trajectories are fully exploited, enabling these two capabilities to co-evolve within a unified algorithmic framework.

\paragraph{Roles and shared policy.}
Our method involves two key roles, the \textbf{extractor} and the \textbf{solver}, which share the same policy parameters. The
\textbf{extractor} takes a completed source interaction as input and produces a
compact textual \emph{skill}—a procedural abstraction that encodes reusable
decision rules, workflow steps, and error-handling strategies extracted from
that experience. The \textbf{solver} takes a new task together with an
injected skill and produces an action trajectory.

\paragraph{Rollout structure.}
Each training iteration begins by solving a source task $x^{\mathrm{src}}\sim\mathcal{D}$
without any injected skill. The solver, parameterized by the shared policy $\pi_{\theta}$, interacts with the environment conditioned on $x^{\mathrm{src}}$ and produces a trajectory: $$\tau \sim \pi_{\theta}(\cdot \mid x^{\mathrm{src}}).$$ The resulting trajectory $\tau$ and environment
reward $r^{\mathrm{src}}$ form the extraction state $s^{e}=(x^{\mathrm{src}},\tau,r^{\mathrm{src}})$, which
conditions the extractor to produce $N$ candidate skills.  A retriever
$\mathcal{R}$ then fetches $K$ semantically related downstream tasks, and for
every (skill, task) pair the solver rolls out a skill-conditioned trajectory.
This structure—source solving $\to$ skill extraction $\to$ downstream
evaluation—produces two coupled sample families within a single iteration:
extractor samples and solver trajectories, both of which are consumed by the
joint objective.

\subsection{Extractor Training}
\label{sec:extractor}

\paragraph{Online skill generation.}
In Evolving-RL, the extractor is optimized in a fully online manner, such that each training iteration begins with the generation of new experience. Given the extraction state $s^{e}$, the extractor samples a set of $N$ candidate skills:
\[
\mathcal{E} = \{e_1,\dots,e_N\},
\qquad
e_i \sim \pi_{\theta}(\cdot \mid s^{e}).
\]
Each candidate $e_i$ is instantiated as a skill, a structured textual abstraction that encodes both high-level procedural knowledge and concrete operational details distilled from the source interaction. The resulting $N$ candidates constitute the comparison set for GRPO-based advantage estimation in extractor optimization.

\paragraph{Generalization-Oriented Evaluation.}
We evaluate the quality and generalization of a skill by its ability to improve solver performance on related tasks. To ensure fair evaluation, all candidate skills within the same comparison group must be assessed on an identical set of downstream tasks. Therefore, we do not use the skill content itself as the retrieval query, as this would cause different skills to be evaluated on different sets of downstream tasks. Instead, downstream tasks are retrieved based on the embedding of the task description.

Concretely, given a source task $x^{\mathrm{src}}$, the retrieval function $R$ returns a set of top-$K$ semantically similar downstream tasks:
\begin{equation}
\mathcal{X}^{\mathrm{ret}}
=
R(x^{\mathrm{src}})
=
\operatorname{TopK}_{x \in \mathcal{D}}
\, s\left(\phi(x^{\mathrm{src}}), \phi(x)\right),
\label{eq:retrieval}
\end{equation}
where $\phi(\cdot)$ denotes the embedding of the task description and $s(\cdot,\cdot)$ is instantiated as cosine similarity. We additionally include the source task itself in $\mathcal{X}^{\mathrm{ret}}$, since skill quality should reflect not only transfer to related tasks but also its effect on the original task.

For each candidate skill $e_i$ and each retrieved downstream task
$x_j \in \mathcal{X}^{\mathrm{ret}}$, the solver generates a skill-conditioned trajectory
$t_{ij}\sim\pi_{\theta}(\cdot\mid x_j,e_i)$ and receives environment reward
$r_{ij}$. The reward assigned to skill $e_i$ is defined as its mean downstream performance:
\begin{equation}
R_i^{e}=\frac{1}{K}\sum_{j=1}^{K} r_{ij}.
\label{eq:skill_reward}
\end{equation}
A skill receives a high reward only if it \emph{consistently} improves solver performance across multiple related tasks, thereby directly aligning the extraction objective with cross-task transfer utility.

\paragraph{Extractor Objective.}
The $N$ candidate skills derived from the same source interaction constitute a single GRPO comparison group. We compute the group-normalized advantage as $A_i^{e}=\frac{R_i^{e}-\operatorname{mean}\left(\{R_{i'}^{e}\}_{i'=1}^{N}\right)}{\operatorname{std}\left(\{R_{i'}^{e}\}_{i'=1}^{N}\right)}$. The extractor loss is
\begin{equation}
\mathcal{L}_{e}(\theta)
=
-\frac{1}{N}\sum_{i=1}^{N}
\min\bigl(
\rho_i^{e}A_i^{e},\;
\operatorname{clip}(\rho_i^{e},1-\epsilon,1+\epsilon)A_i^{e}
\bigr)
+\beta_{e}\,D_{\mathrm{KL}}(\pi_{\theta}\|\pi_{\mathrm{ref}})
-\eta_{e}\,\mathcal{H}(\pi_{\theta}),
\label{eq:extractor_loss}
\end{equation}
where $\rho_i^{e}=\frac{\pi_{\theta}(e_i\mid s^{e})}{\pi_{\theta_{\mathrm{old}}}(e_i\mid s^{e})}$ is the importance sampling ratio, $D_{\mathrm{KL}}$ denotes the KL regularization term, and $\mathcal{H}(\pi_{\theta})$ denotes the entropy term. Contrary to its standard use for encouraging exploration, we employ entropy regularization with a negative coefficient $\eta_e$ to penalize entropy growth and bias the extraction policy toward deterministic predictions. This constraint is necessary because skills naturally admit diverse representations, meaning there is no single deterministic target for extraction. This characteristic renders the process inherently high-entropy. Compounded by the noise introduced during skill evaluation (see Appendix~\ref{app:a1}), these factors could otherwise cause the policy entropy to grow continuously during training, ultimately resulting in training collapse.

\subsection{Solver Training}
\label{sec:solver}

\paragraph{Cross-Skill Group Advantage Estimation.}
During downstream evaluation, each retrieved task $x_j$ yields a set of $N$ trajectories $\{t_{ij}\}_{i=1}^{N}$, where each trajectory is conditioned on a distinct candidate skill. While these trajectories arise from heterogeneous prompts owing to the injected skills, they remain directly comparable, as the underlying task semantics, objective, and contextual structure are shared across all $N$ rollouts for the same task $x_j$. We accordingly aggregate the $N$ skill-conditioned trajectories of each task into a single GRPO comparison group and compute the solver-side advantage through within-group reward normalization:
\begin{equation}
A_{ij}^{s}
=
\frac{%
  r_{ij}-\operatorname{mean}\left(\{r_{i'j}\}_{i'=1}^{N}\right)
}{%
  \operatorname{std}\left(\{r_{i'j}\}_{i'=1}^{N}\right)
}.
\label{eq:solver_advantage}
\end{equation}

Under this grouping mechanism, the correctness of the solver's response on the downstream task emerges as the dominant optimization signal. In the presence of a beneficial skill, the solver is rewarded for exploiting it to surpass trajectories conditioned on suboptimal skills for the same task. Conversely, when provided with spurious or misleading skills, the solver incurs a penalty for deviating from optimal behavior relative to its alternatives, thereby learning to remain robust against degraded guidance. Together, these dual incentives foster a solver that can effectively capitalize on high-quality skills while maintaining stable performance under poor skill conditioning.

\paragraph{Solver objective.}
We reuse all $N\times K$ skill-conditioned trajectories collected during
downstream evaluation.  Let $A_{ij}^{s}$ be the heterogeneous-context advantage
from Eq.~\eqref{eq:solver_advantage}.  The solver loss is
\begin{equation}
\mathcal{L}_{s}(\theta)
=
-\frac{1}{NK}\sum_{j=1}^{K}\sum_{i=1}^{N}
\min\bigl(
\rho_{ij}^{s}A_{ij}^{s},\;
\operatorname{clip}(\rho_{ij}^{s},1-\epsilon,1+\epsilon)A_{ij}^{s}
\bigr)
+\beta_{s}\,D_{\mathrm{KL}}(\pi_{\theta}\|\pi_{\mathrm{ref}}),
\label{eq:solver_loss}
\end{equation}
where $\rho_{ij}^{s}=\frac{\pi_{\theta}(t_{ij}\mid x_j,e_i)}{\pi_{\theta_{\mathrm{old}}}(t_{ij}\mid x_j,e_i)}$, and
$D_{\mathrm{KL}}(\pi_{\theta}\|\pi_{\mathrm{ref}})$ is the KL-divergence regularization
term measuring the deviation of the current policy $\pi_{\theta}$ from the
reference policy $\pi_{\mathrm{ref}}$.

\subsection{Joint Objective and Co-Evolution Dynamics}
\label{sec:joint}

The final training objective is a weighted combination of the extractor and
solver losses:
\begin{equation}
\mathcal{L}
=
\lambda_{e}\mathcal{L}_{e}
+
\lambda_{s}\mathcal{L}_{s}.
\label{eq:total_loss}
\end{equation}
Because both losses operate on the same parameter vector $\theta$, they are
tightly coupled: a gradient step that improves extraction quality also updates
the solver weights.

Concretely, the mechanism operates in two directions.  On the extraction side,
the extractor is incentivized to produce skills that are broadly useful —
skills that help the solver on diverse retrieved tasks.  On the utilization side, the solver
is trained under a realistic distribution of skill quality: it encounters both
informative and noisy skills during the same rollout, which mirrors the
conditions it faces at test time and builds robustness accordingly.

\section{Experiments}
\label{sec:exp}

\subsection{Experimental Setup}
\label{sec:exp-setup}
We instantiate all methods with \texttt{Qwen2.5-7B-Instruct} \cite{qwen2024qwen25} as the base model and evaluate them on two benchmarks with explicit task-level splits: ALFWorld \cite{shridhar2021alfworld} and Mind2Web \cite{deng2023mind2web}. We primarily compare our approach against two categories of baselines: prompt-based experience-driven self-evolution methods ExpeL~\cite{zhao2023expel}, Memento~\cite{zhou2025memento} and ReasoningBank~\cite{ouyang2026reasoningbank} and the RL-based method, GRPO~\cite{shao2024deepseekmath} and SkillRL~\cite{xia2026skillrl}. However, since SkillRL relies on a certain amount of cold-start data, we are unable to evaluate it on Mind2Web. To ensure robustness, all reported results are averaged over five runs. We briefly describe the experimental settings below, and defer detailed implementation choices to Appendix~\ref{app:imp}.

\paragraph{ALFWorld.}
ALFWorld is a text-based embodied benchmark for household task completion via multi-turn interaction. To evaluate generalization in test-time self-evolution, we split the original training set into \textit{seen} and \textit{unseen} subsets by task type. During training, both experience extraction and policy optimization are performed exclusively on the \textit{seen} subset. At test time, the agent is additionally allowed to collect skills from the \textit{unseen} training subset. We report \textbf{success rate} as the evaluation metric on ALFWorld.

\paragraph{Mind2Web.}
Mind2Web is a benchmark for grounded web navigation across diverse real-world websites. We train on the official training split (1009 tasks) and report results on its three standard evaluation settings: cross-task (252 tasks), cross-website (177 tasks), and cross-domain (912 tasks). At test time, given the relatively small size of the original training split and its substantial distribution shift from the evaluation settings, we allow the model to generate trajectories and extract skills on the test set. On Mind2Web, we report two metrics: \textbf{action accuracy (Act.\ Acc.)} and \textbf{success rate (SR)}. Action accuracy is defined as the proportion of correctly predicted actions among all executed actions.

\begin{table}[t]
    \caption{Performance on ALFWorld. Rows labeled ``w/ skills'' evaluate each method augmented with its own extracted skills. The best and second-best results in each column are highlighted in \textbf{bold} and \underline{underline}, respectively. $^{*}$ denotes the results replicated from \cite{xia2026skillrl}.}
    
  \label{tab:alfworld-main}
  \centering
  \small
  \setlength{\tabcolsep}{0pt}
  \begin{tabular*}{\textwidth}{@{\extracolsep{\fill}}lccccccccc@{}}
    \toprule
    & \multicolumn{5}{c}{Seen tasks} & \multicolumn{3}{c}{Unseen tasks} & \multicolumn{1}{c}{All} \\
    \cmidrule(lr){2-6}\cmidrule(lr){7-9}\cmidrule(l){10-10}
    Method & Pick & Heat & Cool & Clean & Avg. & Look & Pick2 & Avg. & Overall \\
    \midrule
    Base Model & 75.0 & 13.0 & 60.0 & 57.4 & 51.9 & 18.9 & 36.5 & 27.4 & 45.5 \\
    Base Model (w/ skills) & 74.2 & 28.7 & 56.2 & 45.8 & 50.9 & 16.7 & 36.5 & 26.3 & 44.5 \\
    \midrule
    \multicolumn{10}{@{}l}{\textit{Prompt-based methods}} \\
    ReasoningBank & 75.0 & 21.7 & 47.6 & 45.2 & 47.5 & 11.1 & 35.3 & 22.9 & 41.1 \\
    ExpeL$^{*}$ & 21.0 & 52.0 & 71.0 & 55.0 & 49.5 & 67.0 & 6.0 & 37.4 & 46.3 \\
    Memento & 79.2 & 39.1 & 76.2 & 64.5 & 64.6 & 0.0 & 41.2 & 20.0 & 53.0 \\
    \midrule
    \multicolumn{10}{@{}l}{\textit{RL-based methods}} \\
    GRPO & \underline{96.7} & \textbf{100.0} & \textbf{100.0} & 90.3 & 96.2 & 53.3 & 12.9 & 33.7 & 79.9 \\
    GRPO (w/ skills) & 95.8 & 98.3 & \underline{98.1} & 96.1 & 97.0 & 61.1 & 27.1 & 44.6 & 83.3 \\
    SkillRL & 93.3 & 72.2 & 80.0 & 95.2 & 86.2 & 81.2 & 55.6 & 68.8 & 81.7 \\
    \midrule
    Ours (w/o skills) & 94.2 & \underline{99.1} & \textbf{100.0} & \underline{96.8} & \underline{97.4} & \textbf{100.0} & \underline{61.2} & \underline{81.1} & \underline{93.1} \\
    Ours (w/ skills) & \textbf{97.5} & \textbf{100.0} & \textbf{100.0} & \textbf{97.4} & \textbf{98.6} & \underline{95.6} & \textbf{81.2} & \textbf{88.6} & \textbf{96.0} \\
    \bottomrule
  \end{tabular*}
\end{table}

\begin{table}[t]
    \caption{Results on Mind2Web under the cross-task, cross-website, cross-domain, and overall evaluation settings. We report both action accuracy (Act. Acc.) and success rate (SR), with the best and second-best results marked in \textbf{bold} and \underline{underline}, respectively.}
    \label{tab:mind2web-main}
    \centering
    \small
    \setlength{\tabcolsep}{0pt}
    \begin{tabular*}{\textwidth}{@{\extracolsep{\fill}}lcccccccc@{}}
        \toprule
        & \multicolumn{2}{c}{Cross-task} & \multicolumn{2}{c}{Cross-website} & \multicolumn{2}{c}{Cross-domain} & \multicolumn{2}{c}{Overall} \\
        \cmidrule(lr){2-3}\cmidrule(lr){4-5}\cmidrule(lr){6-7}\cmidrule(l){8-9}
        Method & Act. Acc. & SR & Act. Acc. & SR & Act. Acc. & SR & Act. Acc. & SR \\
        \midrule
        Base Model & 12.57 & 0.24 & 7.95 & 0.00 & 7.91 & 0.44 & 8.79 & 0.34 \\
        Base Model (w/ skills) & 14.92 & 0.40 & 15.01 & 0.00 & 12.68 & 0.50 & 13.41 & 0.42 \\
        \midrule
        ReasoningBank & 11.84 & 0.00 & 17.76 & 0.56 & 12.09 & 0.66 & 12.79 & 0.52 \\
        Memento & 15.27 & 0.40 & 15.46 & 0.56 & 13.13 & 0.88 & 13.84 & 0.75 \\
        GRPO & 28.79 & 1.51 & 24.61 & 0.00 & 20.84 & 1.67 & 22.83 & 1.42 \\
        GRPO (w/ skills) & 28.98 & 0.87 & 26.20 & 0.90 & 20.33 & \underline{1.69} & 22.73 & 1.43 \\
        \midrule
        Ours (w/o skills) & \underline{35.32} & \underline{1.59} & \textbf{35.15} & \textbf{3.39} & \underline{24.67} & 1.21 & \underline{28.05} & \underline{1.57} \\
        Ours (w/ skills) & \textbf{41.99} & \textbf{2.38} & \underline{35.14} & \underline{1.81} & \textbf{26.97} & \textbf{1.91} & \textbf{30.87} & \textbf{1.99} \\
        \bottomrule
    \end{tabular*}
\end{table}

\subsection{Main Results}
\label{sec:main-results}

Tables~\ref{tab:alfworld-main} and~\ref{tab:mind2web-main} report the main results on ALFWorld and Mind2Web, respectively. Across both benchmarks, Evolving-RL consistently outperforms all baselines, validating both of our main claims: it improves \emph{test-time self-evolution} by strengthening the model's ability to extract and leverage experience, and it also serves as an effective \emph{experience-augmented RL} method that improves the underlying policy even without skill injection at evaluation time.

On ALFWorld, Evolving-RL achieves the best overall success rate of 96.0\% with skill injection, substantially outperforming all baselines. The gain is especially large on unseen tasks, where it improves over GRPO (w/ skills) from 44.6 to 88.6. Even without skill injection, our method still reaches 93.1 overall success rate. On Mind2Web, we observe the same qualitative trend in a noisier and more challenging web environment. Evolving-RL consistently improves over GRPO across the cross-task, cross-website, and cross-domain settings. With skill injection, our method achieves a relative improvement in action accuracy over GRPO by 45.9\%, 42.8\%, and 29.4\% on the cross-task, cross-website, and cross-domain settings, respectively. However, we also observe that on the cross-website split, providing skills does not lead to a clear improvement over the no-skill setting. A possible explanation is that the trajectories generated by the agent on cross-website tasks differ substantially from the trajectory distribution seen during training, making it difficult for the extractor to distill useful skills from such trajectories.

Overall, the results show that Evolving-RL improves performance in two complementary ways. First, our method effectively enhances \emph{test-time self-evolution}: injecting skills at inference time leads to clear and consistent performance gains. Second, the strong performance without skill injection indicates that Evolving-RL is also an effective training algorithm for improving policy generalization itself. In other words, repeated exposure to experience-augmented contexts during training enables the model to internalize reusable procedural patterns into its parameters.

\subsection{Ablations and Analysis}
\label{sec:exp-analysis}

In this section, we present a more comprehensive empirical analysis of Evolving-RL, with the goal of answering the following three questions:

\begin{wraptable}[15]{r}{0.48\textwidth}
\vspace{-3mm}
\centering
\caption{Objective ablation on ALFWorld. The co-evolution objective corresponds to the full Evolving-RL training.}
\label{tab:objective-ablation}
\small
\setlength{\tabcolsep}{3pt}
\begin{tabular*}{\linewidth}{@{\extracolsep{\fill}}llccc@{}}
  \toprule
  Objective & Eval. & Seen & Unseen & Overall \\
  \midrule
  Base Model & w/o skills & 51.9 & 27.4 & 45.5 \\
  Base Model & w/ skills & 50.9 & 26.3 & 44.5 \\
  \midrule
  Extractor-only & w/o skills & 62.6 & 28.6 & 53.7 \\
  Extractor-only & w/ skills & 73.7 & 27.6 & 61.7 \\
  \midrule
  Solver-only & w/o skills & 98.2 & 70.3 & 90.9 \\
  Solver-only & w/ skills & 97.8 & 69.7 & 90.4 \\
  \midrule
  Co-evolution & w/o skills & 97.4 & 81.1 & 93.1 \\
  Co-evolution & w/ skills & \textbf{98.6} & \textbf{88.6} & \textbf{96.0} \\
  \bottomrule
\end{tabular*}
\end{wraptable}

\paragraph{Q1: What drives the generalization gains of Evolving-RL?}  In the main experiments, Evolving-RL exhibits strong generalization even without skill injection at test time. This indicates that its performance gains are not solely attributable to test-time evolution; rather, a substantial portion of the improvement is internalized within the policy parameters. To better isolate the source of this gain, we conducted a controlled ablation study, with results shown in Table~\ref{tab:objective-ablation}. Notably, even the \emph{solver-only} objective---which trains the solver using experience-conditioned contexts without co-evolving the extractor---substantially outperforms GRPO, yielding a 73.2\% relative improvement in generalization. This pattern suggests that injected skills serve as a form of structured auxiliary supervision: during RL training, the policy's repeated exposure to procedural guidance allows it to absorb these regularities directly into its weights, underscoring the efficacy of experience-augmented RL for enhancing generalization.

A critical nuance, however, is that under the \emph{solver-only} objective, test-time skill injection yields no further improvement. This occurs because the policy's self-extracted skills are not sufficiently reliable to support direct conditioning---a limitation evidenced by the performance drop of the Base Model when evaluated \emph{w/ skills}. Consequently, when training only optimizes the policy's ability to utilize skill, the model is repeatedly exposed to noisy skill signals. Under this setup, the most effective strategy is to become insensitive to the injected skills rather than rely on them. This explains why the \emph{solver-only} objective still improves generalization through parameter-level internalization, yet does not benefit further from skill injection at evaluation time.

\begin{wrapfigure}{r}{0.5\textwidth}
    \centering
    \includegraphics[width=0.5\textwidth]{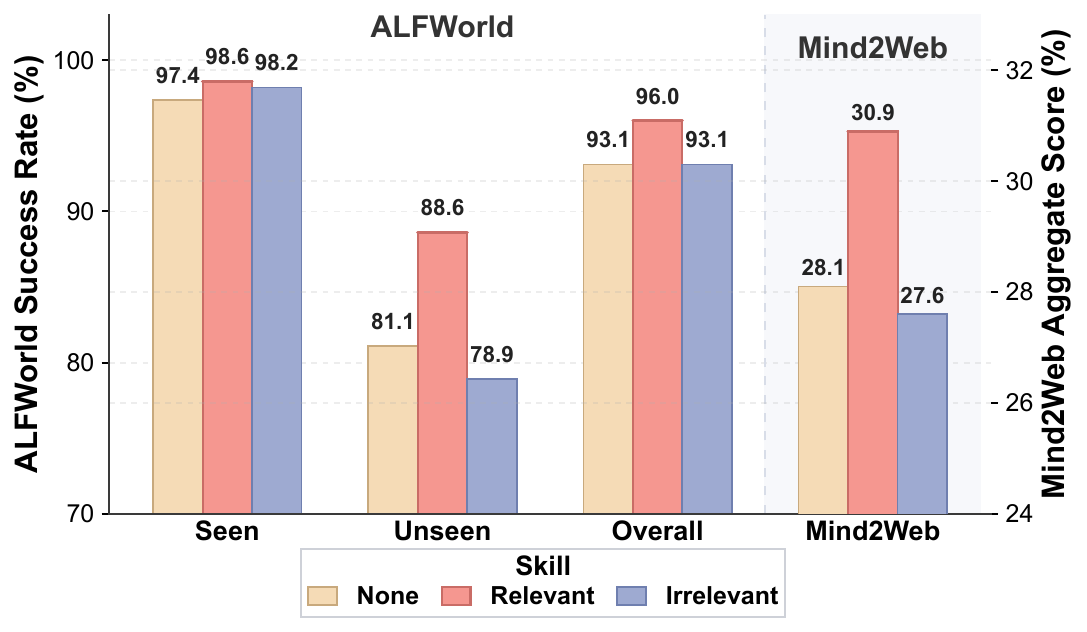}
    \caption{Ablation on skill relevance on ALFWorld and Mind2Web. We compare three settings: no skill injected (\emph{None}), \emph{relevant} skills injected, and \emph{irrelevant} skills injected.}
    \label{fig:re}
\end{wrapfigure}

Conversely, the \emph{extractor-only} objective offers a complementary perspective. Even without explicitly training the solver, the base model's inherent problem-solving capabilities still exhibit measurable improvement, suggesting a fundamental alignment between the parameter optimization directions of the extractor and the solver. However, optimizing the extractor in isolation causes its skill-extraction capability to overfit to \emph{seen} trajectories. As reported in Table~\ref{tab:objective-ablation}, compared with the base model, the extractor-only variant with skills achieves a substantial gain on seen tasks but virtually no improvement on unseen tasks. Considered together with the limitations of the solver-only setup, this overfitting behavior strongly underscores the necessity of co-evolution, which harmonizes the two components to achieve robust and generalizable performance.

\paragraph{Q2: Are the skill-conditioned gains merely due to overfitting to the injected skill context?}

To answer this question, we conduct a controlled ablation study on \emph{skill relevance}. Specifically, we evaluate the model under three conditions: (1) with \emph{relevant} skills, (2) with \emph{irrelevant} skills, and (3) with \emph{no} skill injected. The irrelevant skills are constructed by using the retrieval function $R$ to identify the least relevant questions and then extracting skills from them.

As illustrated in Figure~\ref{fig:re}, injecting irrelevant skills yields performance comparable to the no-skill baseline, whereas providing relevant skills leads to a distinct improvement. This contrast demonstrates that the observed gains are not merely artifacts of overfitting to the presence or format of the context prompts; rather, the policy is genuinely grounded in the semantic \emph{content} of the provided skills. Furthermore, the fact that irrelevant skills do not degrade performance relative to the no-skill setting highlights the robustness of our method: the model can effectively ignore noisy or uninformative signals without suffering a performance penalty.

\paragraph{Q3: Are the extracted skills reusable beyond the training policy?}

We further investigate whether the two capabilities improved by Evolving-RL---skill extraction and skill utilization---are tightly coupled, by examining whether skills extracted by an Evolving-RL-trained policy can transfer to other policies. Specifically, we inject the skills produced by Evolving-RL into both the foundation model and a GRPO-trained model, and additionally measure the number of action steps taken by the agent on successfully completed tasks. Since failed trajectories typically hit the predefined step limit, for a fair comparison we restrict this measurement to 10 successful samples per task type.

\begin{wraptable}[14]{r}{0.55\textwidth}
\vspace{-3mm}
\centering
\caption{Cross-policy transferability of Evolving-RL skills on ALFWorld. We report success rates and average steps for successful tasks. Skills generated by Evolving-RL provide a greater performance boost than the evaluator's self-extracted skills, highlighting strong cross-policy reusability.}

\label{tab:skill-transfer}
\small
\setlength{\tabcolsep}{2pt}
\begin{tabular*}{\linewidth}{@{\extracolsep{\fill}}llcccc@{}}
  \toprule
  Solver & Skill Source & Seen & Unseen & Overall & Steps \\
  \midrule
  Base Model & None & 51.9 & 27.4 & 45.5 & 12.58 \\
  Base Model & Self-extracted & 50.9 & 26.3 & 44.5 & 12.40 \\
  Base Model & Evolving-RL & \textbf{70.3} & \textbf{32.6} & \textbf{60.4} & \textbf{12.20} \\
  \midrule
  GRPO & None & 96.2 & 33.7 & 79.9 & 12.08 \\
  GRPO & Self-extracted & 97.0 & 44.6 & 83.3 & 11.25 \\
  GRPO & Evolving-RL & \textbf{98.0} & \textbf{62.9} & \textbf{88.8} & \textbf{10.04} \\
  \bottomrule
\end{tabular*}
\end{wraptable}

As shown in Table~\ref{tab:skill-transfer}, the skills extracted by Evolving-RL yield substantially larger performance gains than those generated by the evaluation policy itself. Notably, when transferred to the base model, the overall success rate improves markedly from 45.5 to 60.4. This indicates that the skills acquired by Evolving-RL are not narrowly specialized to the training policy, but are broadly reusable across different policies.

Such cross-policy transferability suggests that Evolving-RL enhances the quality of skill extraction itself, rather than merely inducing a private communication protocol between the extractor and the solver. A detailed case study elucidating this improvement is provided in Appendix \ref{app:case}.

\section{Conclusion}
\label{sec:conclusion}
We present \textbf{Evolving-RL}, an end-to-end algorithmic framework for optimizing experience-driven self-evolving capabilities. Evolving-RL conceptualizes the extraction and utilization of experience as a unified process. Anchored in experience extraction and transfer evaluation, it effectively aligns the experience reward with its practical utility, while seamlessly reusing the skill-conditioned trajectories produced during evaluation to jointly train the solver. This design establishes a closed loop in which the quality of experience extraction and the ability to exploit that experience mutually reinforce each other within a single shared policy. In experiments, models trained with Evolving-RL consistently outperform the GRPO baseline in both in-domain performance and out-of-domain generalization. Further analysis suggests that these performance gains arise from both explicit skill conditioning at inference time and the internalization of reusable procedural patterns into the policy parameters, highlighting the dual value of Evolving-RL---both as a means of enhancing self-evolving capabilities and as an experience-augmented RL algorithm. 

\paragraph{Limitations and future work.} 1) Our work primarily focuses on designing a framework to co-optimize experience-driven self-evolution capabilities. Consequently, we employ relatively straightforward strategies for skill management and retrieval. However, we believe that experience evolution mechanisms are crucial not only during the training phase but also for post-deployment adaptation. Therefore, a promising direction for future work is to integrate this framework with more sophisticated evolutionary mechanisms. 2) We theoretically analyzed the estimation error introduced by our evaluation mechanism. Such error can inject noise into training and may even lead to instability in some cases. Reducing this noise and improving the robustness of the evaluation process remains an important direction for future research.

\medskip

{
\small
\bibliographystyle{plainnat}
\bibliography{references}

@article{cai2025flex,
  author = {Cai, Zhicheng and Guo, Xinyuan and Pei, Yu and Feng, Jiangtao and Su, Jinsong and Chen, Jiangjie and Zhang, Ya-Qin and Ma, Wei-Ying and Wang, Mingxuan and Zhou, Hao},
  title = {{FLEX}: Continuous Agent Evolution via Forward Learning from Experience},
  journal = {arXiv preprint arXiv:2511.06449},
  year = {2025},
  doi = {10.48550/arXiv.2511.06449}
}

@article{deng2023mind2web,
  author = {Deng, Xiang and Gu, Yu and Zheng, Boyuan and Chen, Shijie and Stevens, Samuel and Wang, Boshi and Sun, Huan and Su, Yu},
  title = {Mind2Web: Towards a Generalist Agent for the Web},
  journal = {arXiv preprint arXiv:2306.06070},
  year = {2023},
  doi = {10.48550/arXiv.2306.06070}
}

@inproceedings{ouyang2026reasoningbank,
  author = {Ouyang, Siru and Yan, Jun and Hsu, I-Hung and Chen, Yanfei and Jiang, Ke and Wang, Zifeng and Han, Rujun and Le, Long T and Daruki, Samira and Tang, Xiangru and Tirumalashetty, Vishy and Lee, George and Rofouei, Mahsan and Lin, Hangfei and Han, Jiawei and Lee, Chen-Yu and Pfister, Tomas},
  title = {ReasoningBank: Scaling Agent Self-Evolving with Reasoning Memory},
  booktitle = {The Fourteenth International Conference on Learning Representations},
  year = {2026},
  url = {https://openreview.net/forum?id=jL7fwchScm}
}

@article{shridhar2021alfworld,
  author = {Shridhar, Mohit and Yuan, Xingdi and C{\^o}t{\'e}, Marc-Alexandre and Bisk, Yonatan and Trischler, Adam and Hausknecht, Matthew},
  title = {{ALFWorld}: Aligning Text and Embodied Environments for Interactive Learning},
  journal = {arXiv preprint arXiv:2010.03768},
  year = {2021},
  doi = {10.48550/arXiv.2010.03768}
}

@article{xia2026skillrl,
  author = {Xia, Peng and Chen, Jianwen and Wang, Hanyang and Liu, Jiaqi and Zeng, Kaide and Wang, Yu and Han, Siwei and Zhou, Yiyang and Zhao, Xujiang and Chen, Haifeng and Zheng, Zeyu and Xie, Cihang and Yao, Huaxiu},
  title = {SkillRL: Evolving Agents via Recursive Skill-Augmented Reinforcement Learning},
  journal = {arXiv preprint arXiv:2602.08234},
  year = {2026},
  doi = {10.48550/arXiv.2602.08234}
}

@article{zhao2023expel,
  author = {Zhao, Andrew and Huang, Daniel and Xu, Quentin and Lin, Matthieu and Liu, Yong-Jin and Huang, Gao},
  title = {ExpeL: {LLM} Agents Are Experiential Learners},
  journal = {arXiv preprint arXiv:2308.10144},
  year = {2023},
  doi = {10.48550/arXiv.2308.10144}
}

@article{zhou2025memento,
  author = {Zhou, Huichi and Chen, Yihang and Guo, Siyuan and Yan, Xue and Lee, Kin Hei and Wang, Zihan and Lee, Ka Yiu and Zhang, Guchun and Shao, Kun and Yang, Linyi and Wang, Jun},
  title = {Memento: Fine-tuning {LLM} Agents without Fine-tuning {LLM}s},
  journal = {arXiv preprint arXiv:2508.16153},
  year = {2025},
  doi = {10.48550/arXiv.2508.16153}
}

@article{Loshchilov2019Decoupled,
  title   = {Decoupled Weight Decay Regularization},
  author  = {Loshchilov, Ilya and Hutter, Frank},
  journal = {International Conference on Learning Representations (ICLR)},
  year    = {2019}
}

@article{zhang2025qwen3embedding,
  title   = {Qwen3 Embedding: Advancing Text Embedding and Reranking Through Foundation Models},
  author  = {Zhang, Yanzhao and Li, Mingxin and Long, Dingkun and Zhang, Xin and Lin, Huan and Yang, Baosong and Xie, Pengjun and Yang, An and Liu, Dayiheng and Lin, Junyang and Huang, Fei and Zhou, Jingren},
  journal = {arXiv preprint arXiv:2506.05176},
  year    = {2025},
  url     = {https://arxiv.org/abs/2506.05176}
}

@article{qwen2024qwen25,
  title   = {Qwen2.5 Technical Report},
  author  = {Yang, An and Yang, Baosong and Zhang, Beichen and Hui, Binyuan and Zheng, Bo and Yu, Bowen and Li, Chengyuan and Liu, Dayiheng and Huang, Fei and Wei, Haoran and Lin, Huan and Yang, Jian and Tu, Jianhong and Zhang, Jianwei and Yang, Jianxin and Yang, Jiaxi and Zhou, Jingren and Lin, Junyang and Dang, Kai and Lu, Keming and Bao, Keqin and Yang, Kexin and Yu, Le and Li, Mei and Xue, Mingfeng and Zhang, Pei and Zhu, Qin and Men, Rui and Lin, Runji and Li, Tianhao and Tang, Tianyi and Xia, Tingyu and Ren, Xingzhang and Ren, Xuancheng and Fan, Yang and Su, Yang and Zhang, Yichang and Wan, Yu and Liu, Yuqiong and Cui, Zeyu and Zhang, Zhenru and Qiu, Zihan and others},
  journal = {arXiv preprint arXiv:2412.15115},
  year    = {2024},
  url     = {https://arxiv.org/abs/2412.15115}
}

@article{shao2024deepseekmath,
  title   = {DeepSeekMath: Pushing the Limits of Mathematical Reasoning in Open Language Models},
  author  = {Shao, Zhihong and Wang, Peiyi and Zhu, Qihao and Xu, Runxin and Song, Junxiao and Bi, Xiao and Zhang, Haowei and Zhang, Mingchuan and Li, YK and Wu, Y and Guo, Daya and others},
  journal = {arXiv preprint arXiv:2402.03300},
  year    = {2024},
  url     = {https://arxiv.org/abs/2402.03300}
}

@misc{wu2025evolverselfevolvingllmagents,
      title={EvolveR: Self-Evolving LLM Agents through an Experience-Driven Lifecycle}, 
      author={Rong Wu and Xiaoman Wang and Jianbiao Mei and Pinlong Cai and Daocheng Fu and Cheng Yang and Licheng Wen and Xuemeng Yang and Yufan Shen and Yuxin Wang and Botian Shi},
      year={2025},
      eprint={2510.16079},
      archivePrefix={arXiv},
      primaryClass={cs.CL},
      url={https://arxiv.org/abs/2510.16079}, 
}

@inproceedings{huang-chang-2023-towards,
  title = {Towards Reasoning in Large Language Models: A Survey},
  author = {Huang, Jie and Chang, Kevin Chen-Chuan},
  booktitle = {Findings of the Association for Computational Linguistics: ACL 2023},
  pages = {1049--1065},
  year = {2023},
  address = {Toronto, Canada},
  publisher = {Association for Computational Linguistics},
  doi = {10.18653/v1/2023.findings-acl.67},
  url = {https://aclanthology.org/2023.findings-acl.67/}
}

@inproceedings{qiao-etal-2023-reasoning,
  title = {Reasoning with Language Model Prompting: A Survey},
  author = {Qiao, Shuofei and Ou, Yixin and Zhang, Ningyu and Chen, Xiang and Yao, Yunzhi and Deng, Shumin and Tan, Chuanqi and Huang, Fei and Chen, Huajun},
  booktitle = {Proceedings of the 61st Annual Meeting of the Association for Computational Linguistics (Volume 1: Long Papers)},
  pages = {5368--5393},
  year = {2023},
  address = {Toronto, Canada},
  publisher = {Association for Computational Linguistics},
  doi = {10.18653/v1/2023.acl-long.294},
  url = {https://aclanthology.org/2023.acl-long.294/}
}

@article{wang2024survey,
  title = {A survey on large language model based autonomous agents},
  author = {Wang, Lei and Ma, Chen and Feng, Xueyang and Zhang, Zeyu and Yang, Hao and Zhang, Jingsen and Chen, Zhiyuan and Tang, Jiakai and Chen, Xu and Lin, Yankai and Zhao, Wayne Xin and Wei, Zhewei and Wen, Jirong},
  journal = {Frontiers of Computer Science},
  volume = {18},
  number = {6},
  pages = {186345},
  year = {2024},
  doi = {10.1007/s11704-024-40231-1},
  url = {https://link.springer.com/article/10.1007/s11704-024-40231-1}
}

@article{guo2024multiagents,
  title = {Large Language Model based Multi-Agents: A Survey of Progress and Challenges},
  author = {Guo, Taicheng and Chen, Xiuying and Wang, Yaqi and Chang, Ruidi and Pei, Shichao and Chawla, Nitesh V. and Wiest, Olaf and Zhang, Xiangliang},
  journal = {arXiv preprint arXiv:2402.01680},
  year = {2024},
  url = {https://arxiv.org/abs/2402.01680}
}

@inproceedings{wei2022chain,
  title = {Chain-of-Thought Prompting Elicits Reasoning in Large Language Models},
  author = {Wei, Jason and Wang, Xuezhi and Schuurmans, Dale and Bosma, Maarten and Ichter, Brian and Xia, Fei and Chi, Ed and Le, Quoc V. and Zhou, Denny},
  booktitle = {Advances in Neural Information Processing Systems},
  volume = {35},
  pages = {24824--24837},
  year = {2022},
  publisher = {Curran Associates, Inc.},
  url = {https://proceedings.neurips.cc/paper_files/paper/2022/file/9d5609613524ecf4f15af0f7b31abca4-Paper-Conference.pdf}
}

@inproceedings{kojima2022large,
  title = {Large Language Models are Zero-Shot Reasoners},
  author = {Kojima, Takeshi and Gu, Shixiang (Shane) and Reid, Machel and Matsuo, Yutaka and Iwasawa, Yusuke},
  booktitle = {Advances in Neural Information Processing Systems},
  volume = {35},
  pages = {22199--22213},
  year = {2022},
  publisher = {Curran Associates, Inc.},
  url = {https://proceedings.neurips.cc/paper_files/paper/2022/file/8bb0d291acd4acf06ef112099c16f326-Paper-Conference.pdf}
}

@misc{yao2023react,
      title={ReAct: Synergizing Reasoning and Acting in Language Models}, 
      author={Shunyu Yao and Jeffrey Zhao and Dian Yu and Nan Du and Izhak Shafran and Karthik Narasimhan and Yuan Cao},
      year={2023},
      eprint={2210.03629},
      archivePrefix={arXiv},
      primaryClass={cs.CL},
      url={https://arxiv.org/abs/2210.03629}, 
}

@inproceedings{shinn2023reflexion,
  title = {Reflexion: language agents with verbal reinforcement learning},
  author = {Shinn, Noah and Cassano, Federico and Gopinath, Ashwin and Narasimhan, Karthik and Yao, Shunyu},
  booktitle = {Advances in Neural Information Processing Systems},
  volume = {36},
  pages = {8634--8652},
  year = {2023},
  publisher = {Curran Associates, Inc.},
  url = {https://proceedings.neurips.cc/paper_files/paper/2023/file/1b44b878bb782e6954cd888628510e90-Paper-Conference.pdf}
}

@misc{gao2026surveyselfevolvingagentswhat,
      title={A Survey of Self-Evolving Agents: What, When, How, and Where to Evolve on the Path to Artificial Super Intelligence}, 
      author={Huan-ang Gao and Jiayi Geng and Wenyue Hua and Mengkang Hu and Xinzhe Juan and Hongzhang Liu and Shilong Liu and Jiahao Qiu and Xuan Qi and Yiran Wu and Hongru Wang and Han Xiao and Yuhang Zhou and Shaokun Zhang and Jiayi Zhang and Jinyu Xiang and Yixiong Fang and Qiwen Zhao and Dongrui Liu and Qihan Ren and Cheng Qian and Zhenhailong Wang and Minda Hu and Huazheng Wang and Qingyun Wu and Heng Ji and Mengdi Wang},
      year={2026},
      eprint={2507.21046},
      archivePrefix={arXiv},
      primaryClass={cs.AI},
      url={https://arxiv.org/abs/2507.21046}, 
}

@misc{xiang2026systematic,
  title = {A Systematic Survey of Self-Evolving Agents: From Model-Centric to Environment-Driven Co-Evolution},
  author = {Xiang, Zhishang and Yang, Chengyi and Chen, Zerui and Wei, Zhimin and Tang, Yunbo and Teng, Zongpei and Peng, Zexi and Li, Zongxia and Huang, Chengsong and He, Yicheng and Yang, Chang and Wang, Xinrun and Huang, Xiao and Zhang, Qinggang and Su, Jinsong},
  year = {2026},
  month = feb,
  publisher = {TechRxiv},
  doi = {10.36227/techrxiv.177203250.05832634/v2},
  url = {https://doi.org/10.36227/techrxiv.177203250.05832634/v2},
  note = {Preprint}
}

@misc{agrawal2026gepareflectivepromptevolution,
      title={GEPA: Reflective Prompt Evolution Can Outperform Reinforcement Learning}, 
      author={Lakshya A Agrawal and Shangyin Tan and Dilara Soylu and Noah Ziems and Rishi Khare and Krista Opsahl-Ong and Arnav Singhvi and Herumb Shandilya and Michael J Ryan and Meng Jiang and Christopher Potts and Koushik Sen and Alexandros G. Dimakis and Ion Stoica and Dan Klein and Matei Zaharia and Omar Khattab},
      year={2026},
      eprint={2507.19457},
      archivePrefix={arXiv},
      primaryClass={cs.CL},
      url={https://arxiv.org/abs/2507.19457}, 
}

@misc{shao2026agentmisevolveemergentrisks,
      title={Your Agent May Misevolve: Emergent Risks in Self-evolving LLM Agents}, 
      author={Shuai Shao and Qihan Ren and Chen Qian and Boyi Wei and Dadi Guo and Jingyi Yang and Xinhao Song and Linfeng Zhang and Weinan Zhang and Dongrui Liu and Jing Shao},
      year={2026},
      eprint={2509.26354},
      archivePrefix={arXiv},
      primaryClass={cs.AI},
      url={https://arxiv.org/abs/2509.26354}, 
}

@article{brown2020language,
  title={Language models are few-shot learners},
  author={Brown, Tom and Mann, Benjamin and Ryder, Nick and Subbiah, Melanie and Kaplan, Jared D and Dhariwal, Prafulla and Neelakantan, Arvind and Shyam, Pranav and Sastry, Girish and Askell, Amanda and others},
  journal={Advances in neural information processing systems},
  volume={33},
  pages={1877--1901},
  year={2020}
}

@misc{yang2024copsempoweringllm,
      title={CoPS: Empowering LLM Agents with Provable Cross-Task Experience Sharing}, 
      author={Chen Yang and Quanquan Gu and Chenyang Zhao and Dongruo Zhou},
      year={2024},
      eprint={2410.16670},
      archivePrefix={arXiv},
      primaryClass={cs.LG},
      url={https://arxiv.org/abs/2410.16670}, 
}

@misc{wang2024agentworkflowmemory,
      title={Agent Workflow Memory}, 
      author={Zora Zhiruo Wang and Jiayuan Mao and Daniel Fried and Graham Neubig},
      year={2024},
      eprint={2409.07429},
      archivePrefix={arXiv},
      primaryClass={cs.CL},
      url={https://arxiv.org/abs/2409.07429}, 
}

@misc{zheng2025skillweaverwebagentsselfimprove,
      title={SkillWeaver: Web Agents can Self-Improve by Discovering and Honing Skills}, 
      author={Boyuan Zheng and Michael Y. Fatemi and Xiaolong Jin and Zora Zhiruo Wang and Apurva Gandhi and Yueqi Song and Yu Gu and Jayanth Srinivasa and Gaowen Liu and Graham Neubig and Yu Su},
      year={2025},
      eprint={2504.07079},
      archivePrefix={arXiv},
      primaryClass={cs.AI},
      url={https://arxiv.org/abs/2504.07079}, 
}

@misc{cao2026remembermerefineme,
      title={Remember Me, Refine Me: A Dynamic Procedural Memory Framework for Experience-Driven Agent Evolution}, 
      author={Zouying Cao and Jiaji Deng and Li Yu and Weikang Zhou and Zhaoyang Liu and Bolin Ding and Hai Zhao},
      year={2026},
      eprint={2512.10696},
      archivePrefix={arXiv},
      primaryClass={cs.AI},
      url={https://arxiv.org/abs/2512.10696}, 
}

@misc{tang2025agentkbleveraging,
      title={Agent KB: Leveraging Cross-Domain Experience for Agentic Problem Solving},
      author={Xiangru Tang and Ge Zhang and Sirui Hong and Chenglin Wu and Hao Cheng and Jiaheng Liu and Wangchunshu Zhou and Xingyao Wang and He Zhu and Chi Wang and Peng Xia and Daniel Shao and Fang Wu and Xinming Wei and Tianhao Peng and Ziyang Zhou and Tingting Du and Tianrui Qin},
      year={2025},
      eprint={2507.06229},
      archivePrefix={arXiv},
      primaryClass={cs.CL},
      url={https://arxiv.org/abs/2507.06229},
}

@misc{yang2025learningjobexperiencedrivenselfevolving,
      title={Learning on the Job: An Experience-Driven Self-Evolving Agent for Long-Horizon Tasks}, 
      author={Cheng Yang and Xuemeng Yang and Licheng Wen and Daocheng Fu and Jianbiao Mei and Rong Wu and Pinlong Cai and Yufan Shen and Nianchen Deng and Botian Shi and Yu Qiao and Haifeng Li},
      year={2025},
      eprint={2510.08002},
      archivePrefix={arXiv},
      primaryClass={cs.CL},
      url={https://arxiv.org/abs/2510.08002}, 
}

@misc{muhtar2026complementaryreinforcementlearning,
      title={Complementary Reinforcement Learning}, 
      author={Dilxat Muhtar and Jiashun Liu and Wei Gao and Weixun Wang and Shaopan Xiong and Ju Huang and Siran Yang and Wenbo Su and Jiamang Wang and Ling Pan and Bo Zheng},
      year={2026},
      eprint={2603.17621},
      archivePrefix={arXiv},
      primaryClass={cs.LG},
      url={https://arxiv.org/abs/2603.17621}, 
}

@misc{shi2026experientialreinforcementlearning,
      title={Experiential Reinforcement Learning}, 
      author={Taiwei Shi and Sihao Chen and Bowen Jiang and Linxin Song and Longqi Yang and Jieyu Zhao},
      year={2026},
      eprint={2602.13949},
      archivePrefix={arXiv},
      primaryClass={cs.LG},
      url={https://arxiv.org/abs/2602.13949}, 
}

@misc{zhang2026retroagentsolvingevolvingretrospective,
      title={RetroAgent: From Solving to Evolving via Retrospective Dual Intrinsic Feedback}, 
      author={Xiaoying Zhang and Zichen Liu and Yipeng Zhang and Xia Hu and Wenqi Shao},
      year={2026},
      eprint={2603.08561},
      archivePrefix={arXiv},
      primaryClass={cs.AI},
      url={https://arxiv.org/abs/2603.08561}, 
}

@misc{li2026ariseagentreasoningintrinsic,
      title={ARISE: Agent Reasoning with Intrinsic Skill Evolution in Hierarchical Reinforcement Learning}, 
      author={Yu Li and Rui Miao and Zhengling Qi and Tian Lan},
      year={2026},
      eprint={2603.16060},
      archivePrefix={arXiv},
      primaryClass={cs.AI},
      url={https://arxiv.org/abs/2603.16060}, 
}

@misc{alzubi2026evoskillautomatedskilldiscovery,
      title={EvoSkill: Automated Skill Discovery for Multi-Agent Systems}, 
      author={Salaheddin Alzubi and Noah Provenzano and Jaydon Bingham and Weiyuan Chen and Tu Vu},
      year={2026},
      eprint={2603.02766},
      archivePrefix={arXiv},
      primaryClass={cs.AI},
      url={https://arxiv.org/abs/2603.02766}, 
}

@misc{zhou2026mementoskillsletagentsdesign,
      title={Memento-Skills: Let Agents Design Agents}, 
      author={Huichi Zhou and Siyuan Guo and Anjie Liu and Zhongwei Yu and Ziqin Gong and Bowen Zhao and Zhixun Chen and Menglong Zhang and Yihang Chen and Jinsong Li and Runyu Yang and Qiangbin Liu and Xinlei Yu and Jianmin Zhou and Na Wang and Chunyang Sun and Jun Wang},
      year={2026},
      eprint={2603.18743},
      archivePrefix={arXiv},
      primaryClass={cs.AI},
      url={https://arxiv.org/abs/2603.18743}, 
}

@misc{ma2026skillclawletskillsevolve,
      title={SkillClaw: Let Skills Evolve Collectively with Agentic Evolver}, 
      author={Ziyu Ma and Shidong Yang and Yuxiang Ji and Xucong Wang and Yong Wang and Yiming Hu and Tongwen Huang and Xiangxiang Chu},
      year={2026},
      eprint={2604.08377},
      archivePrefix={arXiv},
      primaryClass={cs.AI},
      url={https://arxiv.org/abs/2604.08377}, 
}

@misc{tu2026dynamicdualgranularityskillbank,
      title={Dynamic Dual-Granularity Skill Bank for Agentic RL}, 
      author={Songjun Tu and Chengdong Xu and Qichao Zhang and Yaocheng Zhang and Xiangyuan Lan and Linjing Li and Dongbin Zhao},
      year={2026},
      eprint={2603.28716},
      archivePrefix={arXiv},
      primaryClass={cs.AI},
      url={https://arxiv.org/abs/2603.28716}, 
}

@misc{shen2026skillfoundrybuildingselfevolvingagent,
      title={SKILLFOUNDRY: Building Self-Evolving Agent Skill Libraries from Heterogeneous Scientific Resources}, 
      author={Shuaike Shen and Wenduo Cheng and Mingqian Ma and Alistair Turcan and Martin Jinye Zhang and Jian Ma},
      year={2026},
      eprint={2604.03964},
      archivePrefix={arXiv},
      primaryClass={cs.AI},
      url={https://arxiv.org/abs/2604.03964}, 
}

@misc{jin2025searchr1trainingllmsreason,
      title={Search-R1: Training LLMs to Reason and Leverage Search Engines with Reinforcement Learning}, 
      author={Bowen Jin and Hansi Zeng and Zhenrui Yue and Jinsung Yoon and Sercan Arik and Dong Wang and Hamed Zamani and Jiawei Han},
      year={2025},
      eprint={2503.09516},
      archivePrefix={arXiv},
      primaryClass={cs.CL},
      url={https://arxiv.org/abs/2503.09516}, 
}
}

\newpage
\appendix

\section{Co-Evolution Stability}
\label{app:a1}

\subsection{Reliability of skill evaluation}

Because \(R_i^{e}\) is estimated from a finite number of downstream tasks and
rollouts, we analyze how reliably it can rank two competing candidate skills
\(e_a\) and \(e_b\). For a retrieved task \(x_i\), let
\[
u_{ai}
=
\mathbb{E}\!\left[r(x_i,t)\mid t \sim \pi_{\theta}(\cdot \mid x_i,e_a)\right],
\qquad
u_{bi}
=
\mathbb{E}\!\left[r(x_i,t)\mid t \sim \pi_{\theta}(\cdot \mid x_i,e_b)\right],
\]
denote the true expected rewards of the two skills on task \(x_i\). Their true
average utilities over the \(K\) retrieved tasks are
\[
R_a = \frac{1}{K}\sum_{i=1}^{K} u_{ai},
\qquad
R_b = \frac{1}{K}\sum_{i=1}^{K} u_{bi}.
\]

In our implementation, each skill is evaluated with a single rollout on each
retrieved task. Let
\[
Y_{ai} = r(x_i,t_{ai}), \qquad t_{ai}\sim \pi_{\theta}(\cdot \mid x_i,e_a),
\]
\[
Y_{bi} = r(x_i,t_{bi}), \qquad t_{bi}\sim \pi_{\theta}(\cdot \mid x_i,e_b),
\]
be the corresponding realized rewards. Then the empirical estimates of the two
skills are
\[
\hat R_a = \frac{1}{K}\sum_{i=1}^{K} Y_{ai},
\qquad
\hat R_b = \frac{1}{K}\sum_{i=1}^{K} Y_{bi},
\]
which are unbiased estimators of \(R_a\) and \(R_b\), respectively.

Define the true and empirical performance gaps as
\[
\Delta_{ab} = R_a - R_b,
\qquad
\hat \Delta_{ab} = \hat R_a - \hat R_b.
\]

For each task \(x_i\), the observed rewards \(Y_{ai}\) and \(Y_{bi}\) are
bounded random variables with means \(u_{ai}\) and \(u_{bi}\), respectively.
Although these variables are not necessarily identically distributed across
tasks, they are mutually independent under independent rollout sampling.
Consequently, \(\hat\Delta_{ab}\) is an average of independent, bounded,
non-identically distributed random variables, and a central limit theorem for
such sums implies that its distribution is approximately Gaussian when \(K\) is
moderately large.

Let
\[
v_{ai} = \mathrm{Var}(Y_{ai}),
\qquad
v_{bi} = \mathrm{Var}(Y_{bi}).
\]
Under independent rollout sampling, the variance of the estimated gap is
\begin{equation}
\sigma_{ab}^{2}
=
\mathrm{Var}(\hat\Delta_{ab})
=
\frac{1}{K^{2}}
\sum_{i=1}^{K}
\bigl[
v_{ai} + v_{bi}
\bigr].
\label{eq:sigma_ab}
\end{equation}

Hence, for moderately large \(K\),
\[
\hat\Delta_{ab}
\;\dot\sim\;
\mathcal{N}(\Delta_{ab}, \sigma_{ab}^{2}).
\]
Therefore, if \(R_a > R_b\) (equivalently \(\Delta_{ab}>0\)), the probability
that finite-sample evaluation preserves the correct ordering is approximately
\begin{equation}
\Pr(\hat R_a > \hat R_b \mid R_a > R_b)
\approx
\Phi\!\left(\frac{\Delta_{ab}}{\sigma_{ab}}\right),
\label{eq:ranking_probability_general}
\end{equation}
where \(\Phi(\cdot)\) denotes the cumulative distribution function of the
standard normal distribution.

This expression indicates that the reliability of skill evaluation is jointly
determined by the true performance gap \(\Delta_{ab}\) and the variance
\(\sigma_{ab}^{2}\) of the estimated gap. The larger the reward difference
induced by the two skills, and the smaller the sampling variance, the more
likely the evaluation is to recover their true ordering.

Finally, if rewards are bounded in an interval \([m,M]\), then
\[
v_{ai} \le \frac{(M-m)^{2}}{4},
\qquad
v_{bi} \le \frac{(M-m)^{2}}{4},
\]
which gives
\begin{equation}
\sigma_{ab}^{2}
\le
\frac{(M-m)^{2}}{2K}.
\label{eq:sigma_ab_bound}
\end{equation}
Substituting this into Equation~\eqref{eq:ranking_probability_general} yields
the conservative lower bound
\begin{equation}
\Pr(\hat R_a > \hat R_b \mid R_a > R_b)
\gtrsim
\Phi\!\left(
\frac{\Delta_{ab}\sqrt{2K}}{M-m}
\right).
\label{eq:ranking_probability_bound}
\end{equation}

The Bernoulli-reward case is a special instance of this formulation. When
\(r\in\{0,1\}\), we have \(u_{ai}=p_{ai}\) and
\(v_{ai}=p_{ai}(1-p_{ai})\), which recovers the binary-reward derivation used
for ALFWorld.

\begin{figure}[t]
    \centering
    \includegraphics[width=\linewidth]{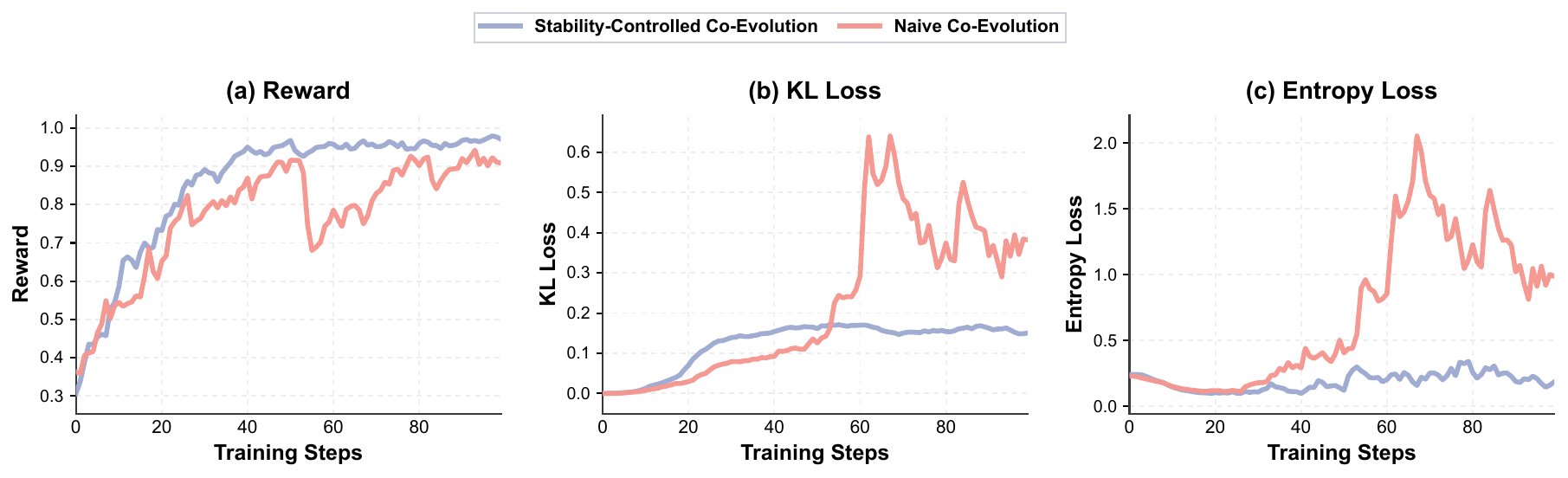}
    \caption{Training stability comparison between stability-controlled co-evolution and naive co-evolution. While naive co-evolution exhibits rapid growth in KL loss and entropy loss, stability-controlled co-evolution keeps both quantities bounded and yields smoother, more reliable reward improvement.}
    
    \label{fig:stability}
\end{figure}

\subsection{Training Stability}
As indicated by Equation~\ref{eq:ranking_probability_general}, the reliability of skill evaluation depends critically on the true performance gap $\Delta_{ab}$ between candidate skills: the larger the gap, the more accurately the evaluation can recover their relative quality. However, RL training involves extensive sampling, and the extractor may occasionally generate rare expressions or anomalous characters. Since textual skill extraction inherently lacks a deterministic ground truth and is fundamentally a high-entropy generation process, the likelihood of producing such irregular outputs is significantly amplified. Importantly, these anomalies do not always cause a substantial drop in downstream solver performance. As a result, the performance gap $\Delta_{ab}$ between a malformed skill and a semantically similar but well-formed skill may remain very small.

When this margin is small, our empirical evaluation mechanism may fail to reliably distinguish between the two skills. Consequently, malformed skills can be assigned positive extractor-side advantages $A_i^{e}$ by chance. This creates a pathological feedback loop: low-probability tokens that happen to appear in such skills are repeatedly reinforced, causing the extractor policy entropy $\mathcal{H}(\pi_{\theta})$ to grow over time. Furthermore, since the extractor and the solver operate on highly similar textual contexts within a shared model, this entropy-driven noise easily bleeds from the extraction process into the solver, further exacerbating the overall instability. Once the entropy exceeds a critical threshold, training becomes unstable and may eventually collapse, as illustrated in Figure~\ref{fig:stability}.

To mitigate this issue, we introduce two explicit stability controls. First, we apply a rule-based filter that assigns zero reward to any candidate skill containing abnormal characters. Second, as defined in Equation~\ref{eq:extractor_loss}, we include the entropy regularization term $-\eta_e \mathcal{H}(\pi_{\theta})$ in the extractor objective to prevent the extraction policy from becoming excessively diffuse. Unlike the standard use of entropy regularization to encourage exploration, our goal here is to suppress uncontrolled entropy growth and constrain the extraction distribution. As shown in Figure~\ref{fig:stability}, these two measures together substantially improve the stability of co-evolutionary training.

The first measure can be viewed as a lightweight modification of the reward function: in addition to transfer-based evaluation, we incorporate a small rule-based reward correction. We also experimented with replacing this rule-based filter by an LLM-based assessment of whether the extracted skill is linguistically natural and fluent. This alternative produced stabilizing effects similar to those of the rule-based correction.

\paragraph{Discussion on Co-Evolution Instability.} The severe instability encountered during co-evolution has also been observed in recent concurrent literature~\cite{muhtar2026complementaryreinforcementlearning}.  They encountered similar training collapse and attributed it to inherent parameter conflicts between the extractor and the solver, ultimately opting to decouple the system into two separate models. However, our analysis above reveals that the true bottleneck lies not in parameter interference, but in the pathological feedback loop driven by evaluation noise and unconstrained extraction entropy. By explicitly identifying this root cause and introducing the corresponding stability controls (rule-based gating and negative entropy regularization), Evolving-RL successfully prevents this collapse. This demonstrates that experience extraction and utilization can indeed be smoothly and jointly optimized within a single, unified policy, eliminating the need for a decoupled architecture.

\section{Case Study}
\label{app:case}

\begin{figure}[t]
    \centering
    \includegraphics[width=\linewidth]{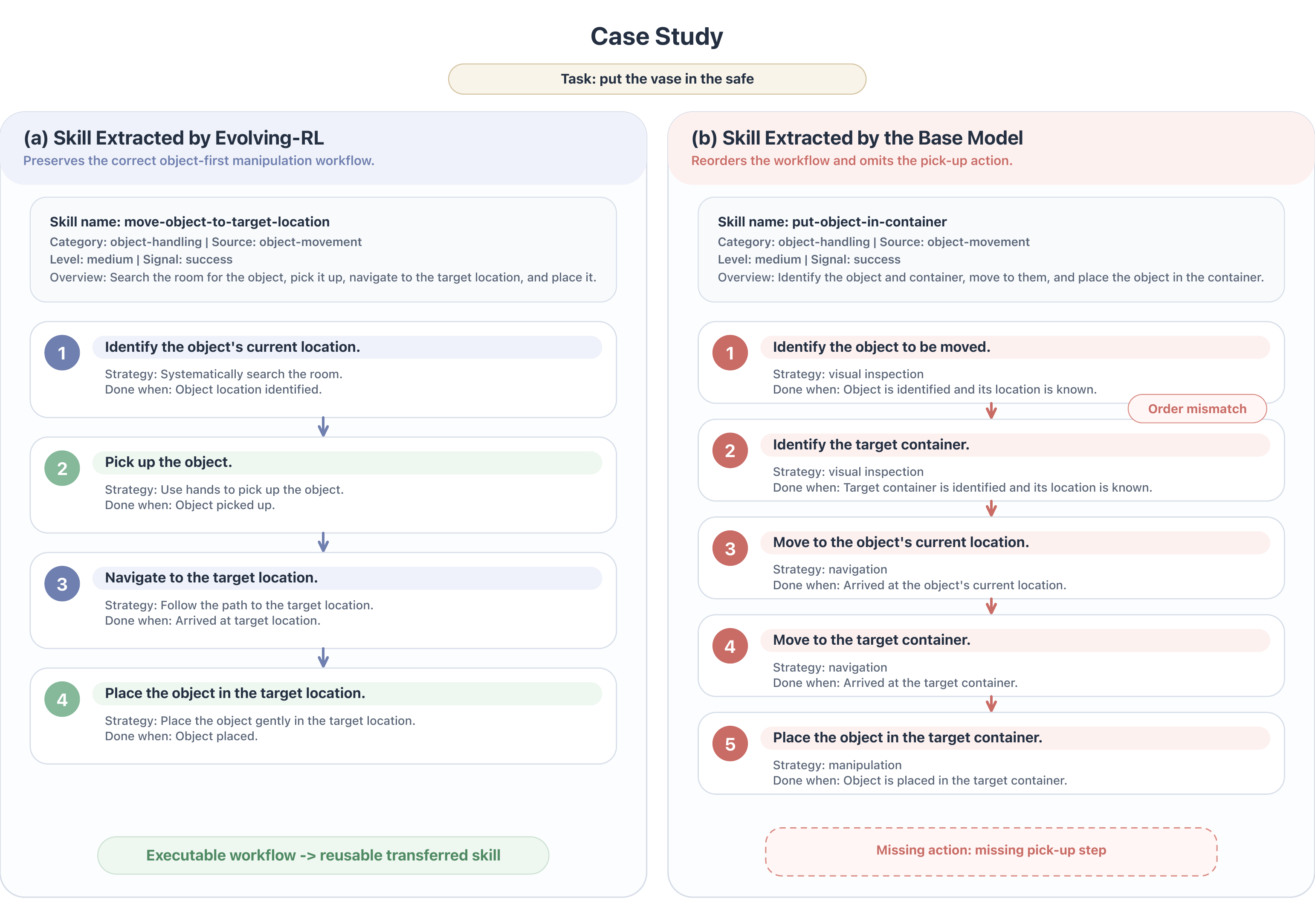}
    \caption{Case study. The skill extracted by the Evolving-RL-trained model is concise and provides a correct procedural guide, whereas the skill extracted by the untrained base model contains misordered steps and omits critical actions.}
    \label{fig:case}
\end{figure}

To better understand how Evolving-RL improves the quality of skill extraction, we compare the skills produced by different models from the same interaction trajectory. As shown in Figure~\ref{fig:case}, for a trajectory corresponding to the task \textit{put the vase in the safe}, the Evolving-RL-trained model extracts a coherent and actionable skill that correctly captures the procedure of locating the target object, picking it up, finding the target container, and placing the object into it. 

In contrast, the skill extracted by the Base Model contains clear procedural flaws. It introduces the step of confirming the container location before moving to the object, resulting in an incorrect ordering of subgoals. Moreover, after locating the object, it fails to include the crucial instruction to \textit{pick up} the object. When such a flawed skill is injected into the solver, it can substantially reduce the solver's success rate.

\section{Implementation Details}
\label{app:imp}

In this section, we describe the experimental setup in detail, including the environment configuration and training implementation.

\subsection{Environment Configuration}

\paragraph{ALFWorld.}
ALFWorld is a text-based interactive environment for embodied household task completion. At each step, the agent receives a textual observation describing the currently visible objects, together with a set of admissible actions. We use the \emph{full interaction history} as the model context, including all previous observations and executed actions. Consequently, each optimization sample in ALFWorld corresponds to a \emph{complete trajectory}. Following \cite{jin2025searchr1trainingllmsreason}, we apply a loss mask to environment observations within the trajectory, such that they are provided as conditioning context but do not contribute to the policy optimization loss.

The environment reward is defined at the trajectory level. If the task is successfully completed, the trajectory receives reward $10$; otherwise, it receives reward $0$. Formally, for a trajectory $\tau$, the reward is
\[
r(\tau)=
\begin{cases}
10, & \text{if the task is completed successfully},\\
0, & \text{otherwise}.
\end{cases}
\]

\paragraph{Mind2Web.}
Mind2Web is a benchmark for grounded web navigation. In the official dataset, each task is paired with a reference action trajectory and the corresponding HTML state at each step. However, raw HTML is prohibitively long and noisy to be directly used as model input. We therefore preprocess each page by filtering the HTML and retaining only interactive elements, such as buttons, input fields, and hyperlinks; the resulting filtered text is used as the agent's observation.

Due to context-length constraints, we adopt a \emph{turn-level} state representation. At each decision step, the agent is given the current filtered observation together with a textual history of previously executed actions, but not the full history of past observations. Therefore, each optimization sample in Mind2Web corresponds to a \emph{single interaction step}, rather than an entire trajectory. 

A key property of Mind2Web is that the dataset only provides the next state along the ground-truth trajectory. If the agent selects an incorrect action at any step, the corresponding next state is unavailable, and the trajectory is terminated immediately. As a result, the agent can proceed to the next decision step only if it predicts the correct action at the current step. Our reported \emph{action accuracy} is computed under this setting, i.e., as the proportion of correctly predicted actions over all executed actions. 

For solver optimization on Mind2Web, since correctness feedback is available at every step, we apply \emph{turn-wise GRPO} to train the solver. Specifically, for the trajectory generated by skill \(e_i\) on retrieved task \(x_j\), the reward at step \(t\) is defined as
\[
r_{ij,t} =
\begin{cases}
1, & \text{if the action predicted at step } t \text{ is correct},\\
0, & \text{otherwise}.
\end{cases}
\]
The extractor reward is computed using \emph{trajectory-level} rewards. Specifically, for the downstream trajectory generated by skill \(e_i\) on retrieved task \(x_j\), its trajectory-level reward is defined as the sum of step-level rewards:

\[
r_{ij}^{\mathrm{traj}} = \sum_{t=1}^{T_{ij}} r_{ij,t},
\]
where $r_{ij,t}$ denotes the reward at step $t$, and $T_{ij}$ is the trajectory length. The extractor reward for skill $e_i$ is then computed as the average trajectory-level reward over the $K$ retrieved downstream tasks:
\[
R_i^{e} = \frac{1}{K}\sum_{j=1}^{K} r_{ij}^{\mathrm{traj}}.
\]
In this way, solver training exploits fine-grained step-level supervision, while extractor evaluation remains aligned with the overall quality of the procedural experience distilled from complete trajectories.

\subsection{Training Setup}

All experiments are initialized from \texttt{Qwen2.5-7B-Instruct} and trained with AdamW \cite{Loshchilov2019Decoupled}. Unless otherwise stated, we use a constant learning rate of $1\times 10^{-6}$, weight decay $0.1$, $\beta_1=0.9$, and $\beta_2=0.98$. Training is performed with GRPO-style policy optimization, using low-variance KL regularization and clipped policy updates. In each rollout, we sample 16 source tasks from the training set, i.e., the rollout batch size is 16. For similar task retrieval, task description embeddings are pre-computed offline using \texttt{Qwen3-Embedding-4B} \cite{zhang2025qwen3embedding} and cached for use during training. All experiments are conducted on 8 NVIDIA H800 GPUs. Each Evolving-RL training run takes approximately 10 hours on ALFWorld and 17 hours on Mind2Web.

Table~\ref{tab:hparam-main} summarizes the main task-specific and method-specific hyperparameters used on ALFWorld and Mind2Web. We train for 75 steps on ALFWorld and 150 steps on Mind2Web. For the GRPO baseline, we use the same number of training steps and match the number of solver samples per update for a fair comparison. Specifically, each GRPO update uses \(B \times N \times K\) samples.

\begin{table}[t]
    \centering
    \caption{Main training hyperparameters for ALFWorld and Mind2Web.}
    \label{tab:hparam-main}
    \small
    \setlength{\tabcolsep}{6pt}
    \begin{tabular}{lcc}
        \toprule
        \textbf{Hyperparameter} & \textbf{ALFWorld} & \textbf{Mind2Web} \\
        \midrule
        Base model & Qwen2.5-7B-Instruct & Qwen2.5-7B-Instruct \\
        Policy clip $\epsilon_{\text{low}}$ & 0.1 & 0.1 \\
        Policy clip $\epsilon_{\text{high}}$ & 0.15 & 0.15 \\
        Number of candidate skills $N$ & 8 & 8 \\
        Number of retrieved tasks $K$ & 4 & 4 \\
        Extractor reward weight $\lambda_e$ & 0.2 & 0.1 \\
        Solver reward weight $\lambda_s$ & 1.0 & 1.0 \\
        Rollout batch size \(B\) & 16 & 16 \\
        Rollout temperature & 1.0 & 1.0 \\
        Rollout top-$p$ & 0.9 & 0.85 \\
        KL loss coefficient & 0.01 & 0.01 \\
        Extractor entropy coefficient & -0.03 & 0.0 \\
        Training steps & 75 & 150 \\
        \bottomrule
    \end{tabular}
\end{table}

Among these hyperparameters, the most important method-specific ones are the number of candidate skills $N$, the number of retrieved downstream tasks $K$, and the extractor/solver reward weights $\lambda_e$ and $\lambda_s$. The parameter $N$ determines the comparison set used for extractor-side GRPO, while $K$ controls how broadly each extracted skill is evaluated for transferability. The weights $\lambda_e$ and $\lambda_s$ balance the contributions of skill extraction and skill utilization during joint optimization.


\end{document}